\documentclass[lettersize,journal]{IEEEtran}
\usepackage{amsmath,amsfonts}
\usepackage{algorithmic}
\usepackage{algorithm}
\usepackage{array}
\usepackage[caption=false,font=normalsize,labelfont=sf,textfont=sf]{subfig}
\usepackage{textcomp}
\usepackage{stfloats}
\usepackage{url}
\usepackage{verbatim}
\usepackage{graphicx}
\usepackage{cite}
\usepackage{algorithm}
\usepackage{algorithmic}

\usepackage{amsfonts}
\usepackage{amsmath}
\usepackage[table]{xcolor}
\usepackage{multirow}
\usepackage{booktabs}
\usepackage{pifont}
\usepackage{fontawesome5}
\usepackage{enumitem}

\begin{document}


\title{Learning Trajectory-Aware Multimodal Large Language Models for
Video Reasoning Segmentation}


\author{
Jingnan Luo$^{1,2}$,
Mingqi Gao$^{1}$,
Jun Liu$^{2}$,
Bin-Bin Gao$^{2}$,
and Feng Zheng$^{1}$\thanks{Corresponding author: Feng Zheng.}
\thanks{$^{1}$Southern University of Science and Technology, Shenzhen, China.}
\thanks{$^{2}$Tencent YouTu Lab, Shenzhen, China.}
\thanks{Jingnan Luo and Mingqi Gao contributed equally to this work.}
}



\maketitle



\begin{abstract}

The prosperity of Multimodal Large Language Models (MLLMs) has stimulated the demand for video reasoning segmentation, which aims to segment video objects based on human instructions. Previous studies rely on unidirectional and implicit text-trajectory alignment, which struggles with trajectory perception when faced with severe video dynamics. In this work, we propose TrajSeg, a simple and unified framework built upon MLLMs. Concretely, we introduce bidirectional text-trajectory alignment, where MLLMs accept grounding-intended (text-to-trajectory) and captioning-intended (trajectory-to-text) instructions. This way, MLLMs can benefit from enhanced correspondence and better perceive object trajectories in videos. The mask generation from trajectories is achieved via a frame-level content integration (FCI) module and a unified mask decoder. The former adapts the MLLM-parsed trajectory-level token to frame-specific information. The latter unifies segmentation for all frames into a single structure, enabling the proposed framework to be simplified and end-to-end trainable. Extensive experiments on referring and reasoning video segmentation datasets demonstrate the effectiveness of TrajSeg, which outperforms all video reasoning segmentation methods on all metrics. The code will be publicly available at https://github.com/haodi19/TrajSeg.

\end{abstract}

\begin{IEEEkeywords}
Video reasoning segmentation, Multimodal Large Language Model.
\end{IEEEkeywords}
\section{Introduction}
\label{sec:intro}

Multimodal Large Language Models (MLLMs)~\cite{liu2024visual,li2025llama,han2024onellm} have revolutionized how we interact with the world, empowering the interpretation of complex visual concepts with human intent. 
Recently, this capability has been pushed into reasoning segmentation~\cite{lai2024lisa}, which aims to predict object masks indicated by human instructions, inspiring considerable follow-up studies~\cite{yang2023improved,pi2024perceptiongpt,ren2024pixellm,he2024multi,xia2024gsva}. 
However, such a surge only occurs in the image domain and faces obstacles when segmenting instruction-relevant video objects, limiting its potential in real-world applications such as interactive video editing and embodied AI. 
    

The gap is mainly raised by two unique natures in videos: (1) Dynamicity and (2) Spatial-temporal consistency. Unlike static image objects, video objects appear in trajectories on multiple frames, with dynamic attributes and behaviours, leading to difficulties in \textit{trajectory perception by human instructions}. In addition, video objects are correlated in space and time, imposing requirements for \textit{mask generation with spatial-temporal coherence}. 

\begin{figure}[!t]
\centering
\includegraphics[width=.98\linewidth]{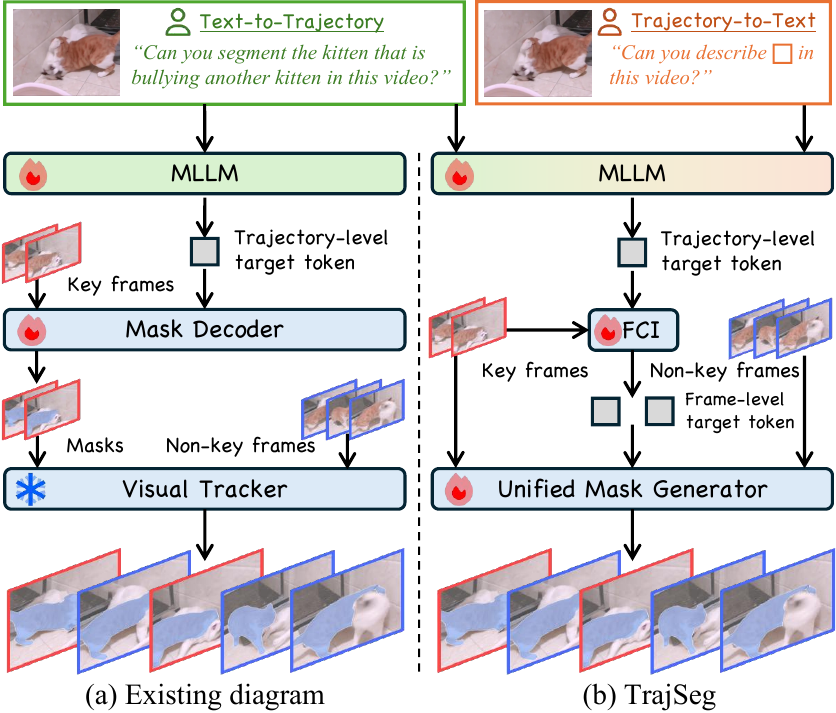}
   \caption{Comparison of existing diagram~\cite{yan2024visa,bai2024one} and ours. (a) learns MLLM via unidirectional alignment (``text-to-trajectory''). Ours considers ``text-to-trajectory'' and ``trajectory-to-text'' to enhance their correspondence (\textcolor{orange}{{\faSquare[regular]}} is the placeholder for trajectory features). Moreover, (b) uses a frame-content integration (FCI) module to refine trajectory tokenization with frame-specific clues. For mask generation, (a) segments key and non-key frames with separately optimized models. (b) supports flexible inputs and unifies all frame segmentation in a single structure, enabling a simplified and end-to-end trainable framework. }
\label{fig:fig1}
\end{figure}

Previous methods~\cite{yan2024visa,bai2024one} achieve video reasoning segmentation by expanding the image-based techniques~\cite{lai2024lisa}. However, they fail to address the above difficulties. First, their trajectory perception is limited due to unidirectional trajectory-text alignment. As shown in Fig.~\ref{fig:fig1} (a), their MLLMs only accept instructions with text-to-trajectory reasoning. This means the alignment can only be achieved implicitly by supervising the final segmentation results. The MLLMs' power in multimodal reasoning remains underutilized. Second, previous methods use the segment-and-track pipeline for efficiency and spatial-temporal consistency. Specifically, they segment key frames upon the trajectory token from MLLMs and then propagate the results to non-key frames via a frozen visual tracker~\cite{cheng2022xmem,bekuzarov2023xmem++}. However, the trajectory-level token lacks frame-specific information and thus limits the segmentation quality. Moreover, Fig.~\ref{fig:fig1} (a) shows that segmentation on key and non-key frames is achieved with separately optimized models, leading to a sub-optimal and complex solution, posing a heavy workload in training and deployment. Therefore, a natural question is raised: \textit{``Can we learn a trajectory-aware and unified model to segment video objects by instructions?''} 

This work answers the question by proposing TrajSeg, an MLLM-driven, \underline{Traj}ectory-aware, and unified framework for video reasoning \underline{Seg}mentation. As shown in Fig.~\ref{fig:fig1} (b), we enable MLLM to support the bidirectional alignment. In addition to the regular grounding-intended instructions (text-to-trajectory), MLLM accepts the captioning-intended ones (trajectory-to-text), where the trajectory is encoded from object-level representations across frames. This enables MLLM to understand the concept and dynamics of trajectories and thus better associate them with human instructions. Furthermore, we propose a novel frame-level content integration (FCI) module, which combines the trajectory-level target token from MLLM with frame-level features, achieving frame-specific embeddings for consistent object segmentation across frames. 

Finally, we propose a unified mask generator to enable TrajSeg to be simplified and end-to-end optimizable. With the unified structure, the generator segments videos with flexible prompts, including target tokens and masks. The former is used for key frames and instruction relevance, and the latter for non-key frames and spatial-temporal consistency. This way, all frames are handled with the same pipeline, as shown in Fig.~\ref{fig:fig1} (b). Compared to previous methods~\cite{yan2024visa,bai2024one}, which rely on separately optimized models, the proposed mask generator calls for less workload in training and deployment. We hope TrajSeg can serve as a strong baseline for video reasoning segmentation. 

The contributions of this work can be summarized as:

\begin{itemize}
\item We propose TrajSeg, an MLLM-driven, trajectory-aware, and unified framework for video reasoning segmentation. Given a video and human instruction, TrajSeg achieves instruction-relevant and spatial-temporal consistent segmentation on all frames in an end-to-end manner. 

\item We present the bidirectional text-trajectory alignment to enhance MLLM's trajectory perception. We further propose a frame content integration (FCI) module to integrate frame-specific clues into MLLM's output, enabling the same object segmentation across frames. 

\item We propose a unified mask generator, which unifies segmentation by instructions and masks into a single structure, benefiting the framework from the simplified training and inference pipeline. 

\item Extensive experiments on both referring and reasoning video segmentation datasets show that TrajSeg outperforms previous video reasoning methods on all metrics. 

\end{itemize}

\section{Related Work}
\label{sec:relate}

\paragraph{Referring Video Object Segmentation (RVOS)}

RVOS aims to segment video objects based on language prompts. This capability of dense vision-language alignment makes RVOS particularly valuable for practical applications such as interactive video editing. Pioneer works~\cite{wu2022language, botach2022end} are based on end-to-end transformers and frame-level multimodal interactions. Subsequent studies improve them in different aspects. For example, SOC~\cite{luo2023soc} and MUTR~\cite{yan2023referred} encode sequence representations to benefit RVOS from sequence-level interactions with texts. In addition, motion properties have been considered in the following works to enhance the dynamic perception, such as HTML~\cite{han2023html}, SgMg~\cite{miao2023spectrum}, Losh~\cite{yuan2024losh}, and DsHmp~\cite{he2024decoupling}. More recently, latent visual-language representation in video diffusion models is explored to leverage large-scale pre-trained generative frameworks to facilitate RVOS~\cite{zhu2024exploring}. 

Recent RVOS studies further explore stronger visual-language grounding and segmentation foundations. For instance, ReferDINO~\cite{liang2025referdino} introduces grounding-guided mask decoding based on visual grounding models, while SSA~\cite{pan2025semantic} improves semantic and sequential alignment between language and video features. 

Despite continuous improvement, RVOS methods lack reasoning abilities in complex contexts and commonsense, limiting their applications. For better performance, previous works select confident masks and use a frozen tracker to propagate them throughout videos. They achieve SoTA scores at the cost of complex and sub-optimal pipelines. This necessitates a unified and end-to-end optimizable framework with strong reasoning abilities and spatial-temporal consistency, which are what we focus on in this work.  

\paragraph{Multimodal Large Language Model (MLLM)}

MLLM aims to perform multimodal tasks upon the strong reasoning abilities of large language models (LLMs)~\cite{yin2023survey}. Pioneering works~\cite{liu2024visual,li2025llama,han2024onellm} leverage the multimodal encoder to bridge with LLMs and respond with texts. More recently, MLLMs have been extended to have multimodal responses by appending extra decoders after LLMs. Specifically, Lai et al.~\cite{lai2024lisa} empowers MLLMs to predict masks by combining with a visual foundation model~\cite{kirillov2023segment}. This gives the birth of reasoning segmentation: the task of segmenting objects by human instructions. Given its huge potential in real-world applications, LISA has inspired many improvements in different aspects. For example, GSVA~\cite{xia2024gsva} and PixelLM~\cite{ren2024pixellm} for multi-mask generation, AnyRef~\cite{he2024multi} for reasoning segmentation with more modalities. 

Recent works also explore leveraging large language models to perform reasoning-oriented video segmentation pipelines. 
For example, AL-Ref-SAM2~\cite{huang2025unleashing} utilizes large language models to perform temporal-spatial reasoning and guide segmentation through foundation models such as GroundingDINO and SAM2.

Although reasoning segmentation has been extended into the video domain~\cite{yan2024visa,bai2024one}, they are directly extended from the image-based techniques~\cite{lai2024lisa} and cannot well handle the unique challenges in video reasoning segmentation. Specifically, they follow an unidirectional text-to-trajectory alignment, which is insufficient in dynamic perception in videos. In addition, as in RVOS methods, previous methods rely on separately optimized models for spatial-temporal consistent segmentation, resulting in a sub-optimal and complex pipeline. In this work, we address the above challenges by proposing a unified framework with bidirectional alignment and an end-to-end trainable pipeline.
\begin{figure*}[!t]
\centering
\includegraphics[width=.8\linewidth]{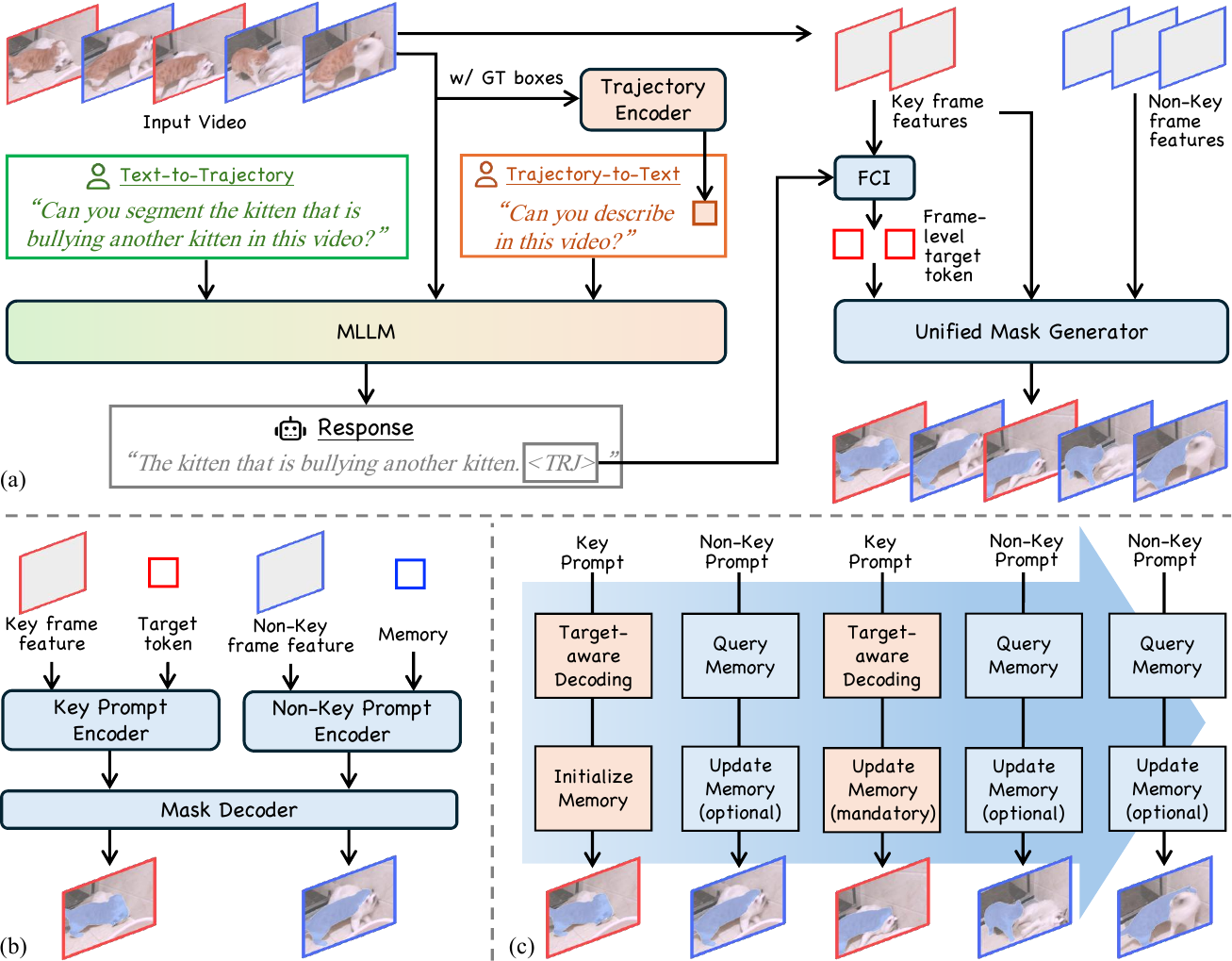}
   \caption{Diagram of TrajSeg. (a) Overall framework; (b) Detailed structure of the unified mask generator; (c) Illustration of how the mask generator processes continuous key \& non-key frames. }
\label{fig:fig2}
\end{figure*}
\section{Method}
\label{sec:method}

\subsection{Overview}

Fig.~\ref{fig:fig2} (a) diagrams TrajSeg, which consists of four modules: (1) MLLM, (2) Trajectory encoder, (3) FCI module, and (4) Unified mask generator. Given a human instruction $\mathcal{I}$ and a video $\mathcal{V}=\{v_t\}^T_{t=1}\in \mathbb{R}^{T\times H\times W\times 3}$ with $T$ frames and each sized of $H\times W$, TrajSeg predicts masks of instruction-referred objects $\mathcal{M}=\{m_t\}^T_{t=1}\in \mathbb{R}^{T\times H\times W}$ on all frames with the same size. 

At first, we uniformly sample $T_\textit{key}$ frames in $\mathcal{V}$, achieving a sub-video $\mathcal{V}_\textit{key}$ capturing the key video context. Then, $\mathcal{V}_\textit{key}$ and $\mathcal{I}$ are fed into MLLM to perceive the special token {\small\textit{$<$TRJ$>$}}, which represents the trajectory-level target information. Concurrently, the visual encoder embeds per-frame visual features $\{f_t\}\in \mathbb{R}^{T\times H/p\times W/p \times C}$ from $\mathcal{V}$, where $p$ is the patch size of the visual backbone. Next, the trajectory-level target token {\small\textit{$<$TRJ$>$}} is enhanced with visual features of key frames $\mathcal{V}_\textit{key}$ via the FCI. Finally, versatile decoder generates masks on all frames, where key frames and others are segmented with the prompts of ``enhanced target tokens'' and ``enhanced target tokens + previous frame masks'', respectively. This way, all frames are segmented under the guidance of instructions while maintaining spatial-temporal consistency with previous frames. The rest of this section presents the main contributions of this work, including bidirectional text-trajectory alignment, FCI, and versatile decoder, followed by the training objectives.

\subsection{Bidirectional Text-Trajectory Alignment}
\label{sec:bid}

Previous methods~\cite{bai2024one, yan2024visa} follow the unidirectional text-to-trajectory alignment. Specifically, their MLLMs receive instructions consisting of textual clues about the target and predict one special token representing target's trajectory across frames. Without extra constraints, the alignment can only be learned implicitly by supervising the segmentation results. Unlike image reasoning, video dynamics further increase the difficulties for alignment, limiting the trajectory perception in previous methods. 

This work presents the bidirectional text-trajectory alignment to enhance MLLM's trajectory perception. As shown in Fig.~\ref{fig:fig2} (a), the MLLM is trained to support instructions with diverse intentions. In particular, besides grounding-intended instructions, it accepts captioning-intended ones, i.e., predicting the textual description from a visual object trajectory. To enable the MLLM to handle trajectories, we devise a trajectory encoder. Given an object trajectory-text pair from the training set, we first encode per-object features $\{f^\textit{obj}_t\}^{N_o}_{t=1}\in \mathbb{R}^{N_o\times 14\times 14\times C}$ by applying ROI-Align on visual features from MLLM, where $N_o$ is the number of frames with objects in the trajectory. Then, we concatenate all object features and use the linear projection to fit the input dimension required by the MLLM, achieving the trajectory feature representation $f_\textit{traj}\in \mathbb{R}^{C}$:


\begin{equation}\label{eq:bta}
f_{\textit{traj}}=\text{Linear}(\text{Concatenate}(f^{\textit{obj}}_1, \cdots, f^{\textit{obj}}_{N_o}))
\end{equation}

Next, the instruction ``\textit{Can you describe $\square$ in this video?}'' is formed, where $\square$ represents $f_\textit{traj}$. Given the instruction, MLLM is trained to predict ``\textit{[Description] $<$TRJ$>$}.'' to convert object trajectory to texts. With textual supervision on generated descriptions, our MLLM learns to better capture the concept of trajectories, thus more clearly understanding trajectory dynamics and spatial-temporal consistency. The knowledge makes the text-to-trajectory correspondence more confident, enabling the MLLM to perceive the target trajectory in videos more effectively by instructions. It is worth noting that the caption-style task in the trajectory-to-text direction is only used during the training stage to enhance the model’s understanding of trajectories. Therefore, no ground-truth information is required during inference.

\subsection{Frame-level Content Integration Module}
\label{sec:fci}
Given the input video and instruction, MLLM predicts one token representing the trajectory of the instruction-relevant targets. Although it is rich in trajectory-level semantics and instruction relevance, it lacks frame-specific spatial information for mask generation. To mitigate this issue, we propose a lightweight and frame-level content integration (FCI) module that expands the trajectory-level target token into the frame-level ones, each containing the spatial information from the corresponding frame. Specifically, given the trajectory-level target token $x_\textit{traj}$ and key frame features $\{f_t\}_{v_t\in \mathcal{V}_\textit{key}}$, FCI module leverages cross-attention to integrate frame-specific information into the target token:  
\begin{equation}\label{eq:fci}
x_\textit{frame}^t=x_\textit{frame}^t+\text{Softmax}\left(\frac{x_\textit{traj}W_Q \cdot (f_tW_K)^T}{\sqrt{d}}\right) \cdot f_tW_V.
\end{equation}

where $W_Q$, $W_k$, and $W_V\in \mathbb{R}^{C\times d}$ are learnable weights. With FCI, the target token can be converted to have not only high-level target semantics but also fine-grained spatial information, leading to more precise mask generation.





\subsection{Unified Mask Generator}
\label{sec:decoder}
Previous referring and reasoning video segmentation methods use a two-stage pipeline to decode instruction-relevant and spatial-temporal consistent masks. As shown in Fig.~\ref{fig:fig1} (a), they first segment key frames with the instruction-aware models and then leverage a frozen visual tracker to propagate masks to non-key frames. However, the segmentation models for key and non-key frames are separately optimized, resulting in a sub-optimal and complex architecture and blocking the mutual interaction between them. 

To address this issue, we propose a unified mask generator that unifies key and non-key frame segmentation into the same structure, achieving instruction relevance and spatial-temporal consistency in an end-to-end manner. As shown in Fig.~\ref{fig:fig2} (b), our generator mainly consists of two prompt encoders and a mask decoder. The former accepts different prompts for key and non-key frames. The latter takes frames and prompts as input and predicts masks: 
\begin{equation}\label{eq:vmd}
m_t = \text{MaskDecoder}(f_t, \text{PromptEncoder}(p_t)),
\end{equation}

\noindent where $p_t$ is the prompt when segmenting the $t^\text{th}$ frame. Therefore, its form depends on the type of video frames. 

For key frames, we set $p_t$ as the frame-level target token, i.e., $p_t=x^t_\textit{frame}$. From Equation~\ref{eq:fci}, it is observed that the target token enhanced from FCI contains the trajectory-level semantics and frame-level spatial details of the target; this allows more precise recovery of the target trajectory on key frames, in both spatial and temporal aspects. For non-key frames, we set $p_t$ as a memory bank by encoding masks from key frames and non-key frames. The memory implicitly represents the target information with sufficient dynamics and spatial-temporal consistency clues. During mask generation, we follow~\cite{ravi2024sam} to segment each non-key frame by querying the memory bank with cross-attention. 

Fig.~\ref{fig:fig2} (c) illustrates how key and non-key frames are decoded with the same structure. At first, we start from one key frame and segment it with the key prompt. Then, the memory bank is initialised and used in subsequent frames. For non-key frames, we only keep recent predictions in the memory for spatial-temporal consistency. For key frames, we keep all their predictions in the memory since they provide instruction-relevant semantics and can be used to refine the memory from error propagation. With this unified structure, the proposed TrajSeg is end-to-end trainable and can utilize more diverse information during decoding, including the overall semantics of the target trajectory, the fine-grained spatial information provided by the FCI module, and even spatial-temporal consistent clues in the memory. Ultimately, we achieve spatio-temporally continuous and instruction-relevant target masks throughout the video.

\subsection{Training Objectives}
\label{sec:train}
The training procedure of TrajSeg is end-to-end and divided into two stages: (1) Pre-training on images and (2) Main-training on videos. The former aims to warm up the model with the fundamental knowledge about reasoning segmentation. The latter fine-tune the model to adapt the knowledge to more challenging video scenarios. In the first stage, the training is performed under the weighted constraint of text loss $\mathcal{L}_{\textit{text}}$ and mask loss $\mathcal{L}_{\textit{mask}}$:
\begin{equation}
\mathcal{L}_\textit{stage1} = \lambda_{\textit{text}} \mathcal{L}_{\textit{text}} + \lambda_{\textit{mask}} \mathcal{L}_{\textit{mask}},
\label{eq:loss1}
\end{equation}
where $\mathcal{L}_{\textit{text}}$ is the auto-regressive cross-entropy (CE) loss for text generation. $\mathcal{L}_{\text{mask}}$ is the combination of per-pixel binary cross-entropy (BCE) loss and DICE loss. 

In the second stage, as some video frames may not contain the target object, a classification loss $\mathcal{L}_\textit{cls}$ is incorporated to indicate the target's presence. During training, we compute $\mathcal{L}_\textit{mask}$ only on frames where the target presents: 
\begin{equation}
\mathcal{L}_\textit{stage2} = \lambda_{\textit{text}} \mathcal{L}_{\textit{text}} + \lambda_{\textit{cls}} \mathcal{L}_{\textit{cls}} + \lambda_{\textit{mask}} \mathcal{L}_{\textit{mask}} \cdot p,
\label{eq:loss2}
\end{equation}

\noindent where $p=1$ when the target presents and $p=0$ otherwise. $\mathcal{L}_{\textit{cls}}$ is the binary cross-entropy (BCE) loss for the target presence score. Overall, all loss terms in Equations~\ref{eq:loss1} and \ref{eq:loss2} are defined as follows:
{\small
\begin{equation}
\begin{aligned}
\mathcal{L}_{\textit{text}}\hspace{4pt}  &= \text{CE}(\hat{y}_{\textit{text}}, y_{\textit{text}}),\ \hspace{4pt}\quad \mathcal{L}_{\textit{cls}} = \lambda_{\textit{cls}} \text{BCE}(\hat{p}, p), \\ 
\mathcal{L}_{\textit{mask}} &= \lambda_{\textit{bce}} \text{BCE}(\hat{\mathcal{M}}, \mathcal{M}) + \lambda_{\textit{dice}} \text{DICE}(\hat{\mathcal{M}}, \mathcal{M}),
\end{aligned}
\end{equation}
}

\noindent where ($\hat y_\textit{text}$, $y_\textit{text}$), ($\hat{\mathcal{M}}$, $\mathcal{M}$), and ($\hat p$, $p$) are ground truth and predicted MLLM response, masks, and target presence.

\section{Experiments}
\label{sec:exp}


\subsection{Experimental Setting}
\label{sec:exp:setting}
\paragraph{Training Data} 
We train TrajSeg on diverse data for better generalization. During per-training, we use image data in semantic segmentation (ADE20K~\cite{zhou2017scene}, COCO-Stuff~\cite{caesar2018coco}, PACO-LVIS~\cite{ramanathan2023paco}, PASCAL-Part~\cite{chen2014detect}), referring segmentation (Ref-COCO~\cite{yu2016modeling}, Ref-CLEF~\cite{kazemzade2014referring}), VQA (LLaVA-Instruct-150k~\cite{liu2023improvedllava}), and reasoning segmentation (ReasonSeg~\cite{lai2024lisa}). During main-training, we use video data in RVOS (Ref-YouTube-VOS~\cite{seo2020urvos} and MeViS~\cite{ding2023mevis}) and reasoning video segmentation data (ReVOS~\cite{yan2024visa}). Additionally, we sample images from Ref-COCO~\cite{yu2016modeling}, Ref-COCO~\cite{yu2016modeling}, and Ref-COCOg~\cite{mao2016generation} to form pseudo-videos to expand training samples. 

To facilitate bi-directional learning and unify mask decoding, we format our video samples into three types: 


\begin{itemize}
  \item \textit{Grounding (Text-to-Trajectory)}: Given an input video and an instruction, TrajSeg predicts a mask trajectory of the instruction-referred object. Therefore, our MLLM's input is formatted as: ``{\textit{Can you segment the [description] in this video?}}''. The expected response is: ``{\textit{Sure, [description] $<$TRJ$>$.}}''
  \item \textit{Captioning (Trajectory-to-Text)}: Given an input video, an instruction, and a trajectory of objects across frames, the goal is to describe the object's trajectory. Therefore, the instruction template is: ``{\textit{Can you describe $\square$ in this video?}}”, where $\square$ represents the encoded features of the trajectory. The response template for the MLLM is the same as for the grounding data.
  \item \textit{Tracking}: Given a video and the first frame mask, the goal is to predict the masks for subsequent frames. This category supports the mask decoder to track with memory. Therefore, the samples do not pass through the MLLM, and the textual cross-entropy (CE) loss is not employed. 
\end{itemize}
\paragraph{Implementation Details} 
We use the pre-trained LLaVA-7B~\cite{liu2024visual} and SAM-2's decoder~\cite{ravi2024sam} to initialize our MLLM and mask generator. The trainable components of our end-to-end framework include the MLLM with LoRA~\cite{hu2021lora}, mask generator, FCI module, and trajectory encoder, while others remain frozen. During training with video data, we sample 10 frames from each video to form the pseudo input. During pre-training, we follow the regular reasoning segmentation pipeline~\cite{lai2024lisa} on image datasets, which lasts for 10 epochs. During main-training, we use all the data for mixed training. In particular, we use 20\% of the video data as tracking-style samples. The remaining 80\% of the data is used for joint training of the MLLM and mask generator, where we sample grounding-style and captioning-style data equally and only consider ReVOS's data for grounding samples. The main-training lasts for 14 epochs. All training runs on 4 GPUs, with a batch size of 80 and the distributed training engine from DeepSpeed~\cite{rasley2020deepspeed}. The AdamW optimizer~\cite{loshchilov2017decoupled} is configured with a learning rate and weight decay of 0.0003 and 0. Prior to input into the MLLM, all images/frames are resized to $224\times 224$. We set weights of the text generation loss $\lambda_{\textit{text}}$ and the $\lambda_{\textit{bce}}$ at 1.0 and 2.0. Additionally, the weights for the DICE loss $\lambda_{\textit{bce}}$ and the object loss $\lambda_{\textit{cls}}$ are set at 0.5.

\paragraph{Evaluation Protocols} We evaluate TrajSeg on video referring and reasoning benchmarks to validate its effectiveness and generalization. For referring segmentation, we consider Ref-YouTube-VOS (Seo et al. 2020), Ref-DAVIS (Khoreva et al. 2018), and MeViS~\cite{ding2023mevis}, the most popular and challenging benchmarks in this field. For video reasoning segmentation, we consider ReVOS~\cite{yan2024visa}, which considers both referring and reasoning data. The evaluation metrics are the Jaccard index $\mathcal{J}$ (regional accuracy), F-measure $\mathcal{F}$ (contour accuracy), and their average $\mathcal{J}\&\mathcal{F}$. All metrics are higher-is-better. 

\subsection{Results}
\label{sec:exp:results}


\paragraph{Video Referring Segmentation} Comparisons in Tab.~\ref{tab:overall_comparison} show that TrajSeg achieves the best overall performance among reasoning-based video segmentation methods across all benchmarks, while remaining competitive with task-specific RVOS architectures.
With the same MLLM base, TrajSeg surpasses TrackGPT~\cite{trackgpt} and VISA~\cite{yan2024visa} by a large margin, with the least advantage of 10.1, 5.9, and 8.4 on $\mathcal{J}\&\mathcal{F}$ on different benchmarks.

We also include several recent RVOS works, such as SSA~\cite{pan2025semantic} and ReferDINO~\cite{liang2025referdino}. These methods mainly improve task-specific RVOS architectures or segmentation backbones for language-conditioned video segmentation. For example, SSA and ReferDINO focus on stronger visual-language grounding and object-level representations. In contrast, TrajSeg focuses on trajectory-aware reasoning representation within an MLLM-driven framework, enabling explicit modeling of object dynamics and unified end-to-end optimization.

On the more challenging MeViS benchmark, TrajSeg surpasses all reasoning-based methods and achieves competitive results compared with recent RVOS architectures.
We conjuncture that MeViS focuses on motion descriptions with complex dynamics of the target object, thus requiring strong capabilities in video understanding and trajectory perception.
Additionally, MeViS videos are far longer than Ref-YouTube-VOS and Ref-DAVIS, posing extra difficulties in modeling trajectories with consistent targets.
With a trajectory-aware and unified framework, our TrajSeg model can better handle both challenges than existing reasoning-based methods.

\paragraph{Video Reasoning Segmentation} 
Comparisons in Tab.~\ref{tab:overall_comparison_reasoning} show that TrajSeg's scores on ReVOS far exceed those of the methods with the same MLLM, even compared to VISA-13B~\cite{yan2024visa}, TrajSeg's performance is on par, showing the effectiveness in video reasoning segmentation.

\begin{figure}[!t]
\centering
\includegraphics[width=\linewidth]{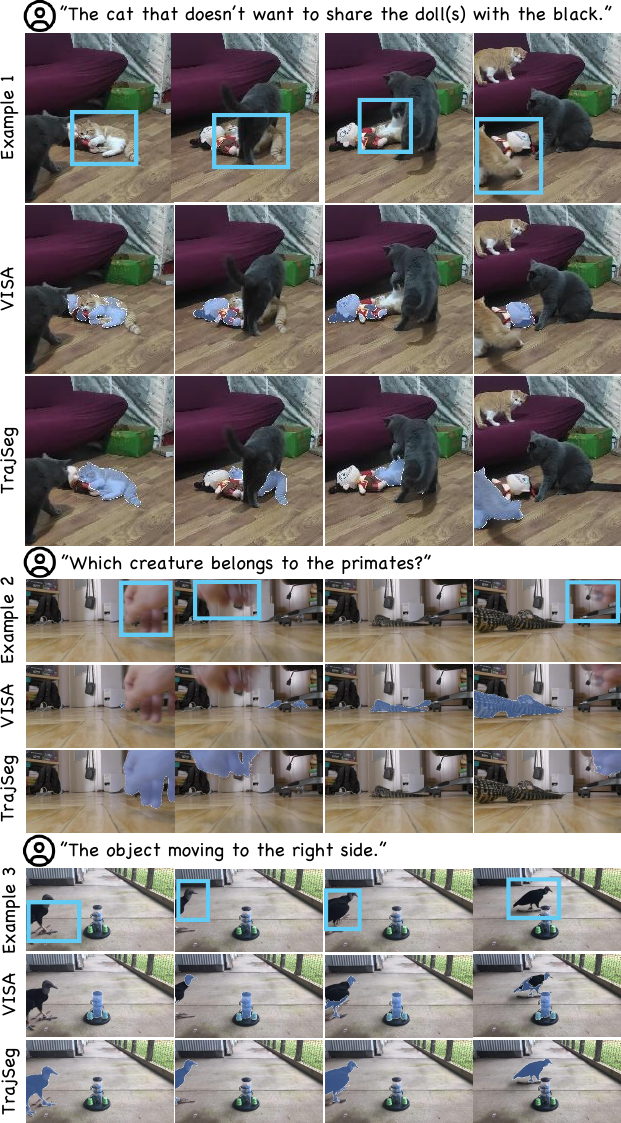}
   \caption{Comparisons of TrajSeg and VISA-7B~\cite{yan2024visa} on ReVOS. Blue boxes are Ground Truth objects.}
\label{fig:visual}
\end{figure}
\paragraph{Qualitative Comparisons }
To show TrajSeg's effectiveness, we visualize our results on ReVOS and compare them with VISA-7B~\cite{yan2024visa}, the only open-sourced work in this field, as shown in Fig.~\ref{fig:visual}. 

\begin{itemize}
  \item Example 1 requires \textit{dynamic reasoning} due to action-dominant instructions. Results show VISA decodes $>1$ objects and fails to predict target trajectories. In contrast, TrajSeg decodes correct trajectories with consistent identity, showing its effective parsing of video dynamics. 
  \item Example 2 requires \textit{commonsense reasoning} due to world knowledge instructions. Results show VISA focuses on salient objects, while TrajSeg correctly identifies the target, showing its strong commonsense reasoning and robustness against salient distractors. 
  \item Example 3 verifies the capability in \textit{recognizing moving objects}. Results show VISA struggles with target trajectory perception, decoding multiple objects with incomplete regions. In contrast, TrajSeg accurately identifies the moving target and predicts precise masks.
\end{itemize}

\newcommand{\jfwidth}{0.105\columnwidth}
\newcommand{\linevspace}{\noalign{\vspace{0.05pt}}}

\begin{table*}[t!]
\footnotesize
\tabcolsep=0.135cm
\centering
\caption{Quantitative comparisons on referring video segmentation benchmarks (Ref-YouTube-VOS, Ref-DAVIS, and MeViS). }
\label{tab:overall_comparison}
\vspace{-0.6em}
  \begin{tabular}{p{0.5\columnwidth} |  p{0.29\columnwidth}<{\centering} | 
  p{\jfwidth}<{\centering} p{\jfwidth}<{\centering}
  p{\jfwidth}<{\centering} | p{\jfwidth}<{\centering}
  p{\jfwidth}<{\centering} p{\jfwidth}<{\centering} |
  p{\jfwidth}<{\centering} p{\jfwidth}<{\centering}
  p{\jfwidth}<{\centering} }
  \toprule
     \multirow{2}{*}{Method} & \multirow{2}{*}{Backbone} & \multicolumn{3}{c|}{Ref-YouTube-VOS} & \multicolumn{3}{c|}{Ref-DAVIS} & \multicolumn{3}{c}{MeViS} \\
     \linevspace
     \cline{3-11}
     \linevspace
     & & $\mathcal{J}\&\mathcal{F}$ & $\mathcal{J}$ & $\mathcal{F}$ & $\mathcal{J}\&\mathcal{F}$ & $\mathcal{J}$ & $\mathcal{F}$ & $\mathcal{J}\&\mathcal{F}$ & $\mathcal{J}$ & $\mathcal{F}$ \\
    \linevspace
    \hline

    \rowcolor{gray!17!white}
    \multicolumn{11}{c}{\textit{Referring video segmentation methods}} \\
    \hline
    \linevspace

    MTTR~\cite{botach2022end} & Video-Swin-T & 55.3 & 54.0 & 56.6 & - & - & - & 30.0 & 28.8 & 31.2 \\

    ReferFormer~\cite{wu2022language} & Video-Swin-B & 62.9 & 61.3 & 64.6 & 61.1 & 58.1 & 64.1 & 31.0 & 29.8 & 32.2 \\

    LMPM~\cite{ding2023mevis} & Swin-T & - & - & - & - & - & - & 37.2 & 34.2 & 40.2 \\

    OnlineRefer~\cite{wu2023onlinerefer} & Swin-L & 63.5 & 61.6  & 65.5 & 64.8 & 61.6 & 67.7 & - & - & - \\

    VD-IT~\cite{zhu2024exploring} & Video Diffusion & 66.5 & 64.4 & 68.5 & 69.4 & 66.2 & 72.6 & - & - & - \\

    DsHmp~\cite{he2024decoupling} & Video-Swin-B & 67.1 & 65.0 & 69.1 & 64.9 & 61.7 & 68.1 & 46.4& \underline{43.0} & \underline{49.8}\\

    SSA~\cite{pan2025semantic} & CLIP-B & 64.3 & 62.2 & 66.4 & 67.3 & 64.0 & 70.7 & 48.6 & 44.0 & 53.2 \\

    ReferDINO~\cite{liang2025referdino} & GroundingDINO-B & \textbf{69.3} & \textbf{67.0} & \textbf{71.5} & 68.9 & 65.1 & \underline{72.9} & \textbf{49.3} & 44.7 & \textbf{53.9} \\

    \linevspace
    \hline
    
    \rowcolor{gray!17!white}
    \multicolumn{11}{c}{\textit{Reasoning video segmentation methods}} \\

    \hline
    \linevspace

    TrackGPT~\cite{trackgpt} & LLaVA-7B & 56.4 & 55.3 & 57.4 & 63.2 & 59.4 & 67.0 & 40.1 & 37.6 & 42.6\\

    TrackGPT~\cite{trackgpt} & LLaVA-13B & 59.5 & 58.1 & 60.8 & 66.5 & 62.7 & 70.4 & 41.2 & 39.2 & 43.1 \\

    VISA~\cite{yan2024visa} & LLaVA-7B & 53.9 & 53.4 & 54.3 & 64.8 & 62.2 & 67.3 & 37.2 & 35.1 & 39.4 \\

    VISA~\cite{yan2024visa} & LLaVA-13B & 54.4 & 54.0 & 54.8 & 66.0 & 63.2 & 68.8 & 37.9 & 35.8 & 40.0 \\

    VISA~\cite{yan2024visa} & Chat-UniVi-7B & 61.5 & 59.8 & 63.2 & 69.4 & 66.3 & 72.5 & 43.5 & 40.7 & 46.3 \\

    VISA~\cite{yan2024visa} & Chat-UniVi-13B & 63.0 & 61.4 & 64.7 & 70.4 & 67.0 & 73.8 & 44.5 & 41.8 & 47.1 \\

    VideoLISA~\cite{bai2024one} & LLaVA-Phi-3.8B & 61.7 & 60.2 & 63.3 & 67.7 & 63.8 & 71.5 & 42.3 & 39.4 & 45.2\\ 

    VideoLISA (p)~\cite{bai2024one} & LLaVA-Phi-3.8B & 63.7 & 61.7 & 65.7 & 68.8 & 64.9 & 72.7 & 44.4 & 41.3 & 47.6 \\ 

    AL-Ref-SAM2~\cite{huang2025unleashing} & GPT-4 & \underline{67.9} & \underline{65.9} & \underline{69.9} & \textbf{74.2} & \textbf{70.4} & \textbf{78.0} & 42.8 & 39.5 & 46.2 \\
    
    \rowcolor{blue!17!white}
    Ours & LLaVA-7B & 67.0 & 65.2 & 68.7 & \underline{69.6} & \underline{66.4} & \underline{72.9} & \underline{48.7} & \textbf{45.8} & 51.6\\

    \bottomrule
  \end{tabular}
  
\end{table*}

\newcommand{\jfwidthrevos}{0.105\columnwidth}
\begin{table*}[t!]
\footnotesize
\tabcolsep=0.135cm
\centering
\caption{Quantitative comparisons on reasoning video segmentation benchmarks (ReVOS). }
\label{tab:overall_comparison_reasoning}
\vspace{-0.6em}
  \begin{tabular}{p{0.5\columnwidth} | p{0.29\columnwidth}<{\centering} | 
  p{\jfwidthrevos}<{\centering} p{\jfwidthrevos}<{\centering}
  p{\jfwidthrevos}<{\centering} | p{\jfwidthrevos}<{\centering}
  p{\jfwidthrevos}<{\centering} p{\jfwidthrevos}<{\centering} | p{\jfwidthrevos}<{\centering}
  p{\jfwidthrevos}<{\centering} p{\jfwidthrevos}<{\centering}}
  \toprule
     \multirow{2}{*}{Method} & \multirow{2}{*}{Backbone} & \multicolumn{3}{c|}{Referring} & \multicolumn{3}{c|}{Reasoning} & \multicolumn{3}{c}{Overall} \\
     \linevspace
     \cline{3-11}
     \linevspace
     & & $\mathcal{J}\&\mathcal{F}$ & $\mathcal{J}$ & $\mathcal{F}$ & $\mathcal{J}\&\mathcal{F}$ & $\mathcal{J}$ & $\mathcal{F}$ & $\mathcal{J}\&\mathcal{F}$ & $\mathcal{J}$ & $\mathcal{F}$ \\
    \linevspace
    \hline

    \rowcolor{gray!17!white}
    \multicolumn{11}{c}{\textit{Referring video segmentation methods}} \\

    \hline
    \linevspace



    MTTR~\cite{botach2022end} & Video-Swin-T  & 30.0 & 29.8 & 30.2 & 21.0 & 20.4 & 21.5 &  25.5 & 25.1 & 25.9 \\

    ReferFormer~\cite{wu2022language} & Video-Swin-B  & 32.7 & 31.2 & 34.3 & 23.4 & 21.3 & 25.6 & 28.1 & 26.2 & 29.9 \\

    LMPM~\cite{ding2023mevis} & Swin-T  & 34.1 & 29.0 & 39.1 & 18.8 & 13.3 & 24.3 & 26.4 & 21.2 & 31.7 \\




    \linevspace
    \hline
    
    \rowcolor{gray!17!white}
    \multicolumn{11}{c}{\textit{Reasoning video segmentation methods}} \\

    \hline
    \linevspace

    TrackGPT~\cite{trackgpt} & LLaVA-7B  & 48.2 & 46.7 & 49.7 & 39.0 & 36.8 & 41.2 & 43.6 & 41.8 & 45.5 \\

    TrackGPT~\cite{trackgpt} & LLaVA-13B & 49.5 & 48.3 & 50.6 & 40.5 & 38.1 & 42.9 & 45.0 & 43.2 & 46.8 \\

    VISA~\cite{yan2024visa} & LLaVA-7B  & 51.0 & 49.4 & 52.6 & 43.2 & 40.5 & 45.8 & 47.1 & 44.9 & 49.2 \\

    VISA~\cite{yan2024visa} & LLaVA-13B  & \textbf{57.4} & \textbf{55.7} & \underline{59.0} & 44.2 & 41.9 & 46.5 & 50.8 & \textbf{48.8} & 52.8 \\

    VISA~\cite{yan2024visa} & Chat-UniVi-7B  & 50.9 & 49.2 & 52.6 & 43.0 & 40.6 & 45.4 & 46.9 & 44.9 & 49.0 \\

    VISA~\cite{yan2024visa} & Chat-UniVi-13B  & \textbf{57.4} & \underline{55.6} & \textbf{59.1} & \underline{44.3} & \underline{42.0} & \underline{46.7} & \underline{50.9} & \textbf{48.8} & \underline{52.9} \\

    \rowcolor{blue!17!white}
    Ours & LLaVA-7B & 56.0 & 53.9 & 58.1 & \textbf{46.1} & \textbf{43.7} & \textbf{48.5} & \textbf{51.1} & \textbf{48.8} & \textbf{53.3} \\
         
    \bottomrule
  \end{tabular}

\vspace{-1em}
\end{table*}

\subsection{Analysis}
\label{sec:exp:ablations}

We ablate TrajSeg on ReVOS~\cite{yan2024visa} since it has referring and reasoning data. 

\newcommand{\jfwidthabl}{0.087\columnwidth}
\paragraph{Bidirectional Alignment}
Tab.~\ref{tab:abl_both} shows the impact of bidirectional alignment on TrajSeg. For fair and efficient comparison, we build two TrajSeg variants w/ and w/o the captioning task and train them for 10 epochs. It is observed that bidirectional alignment works better on the reasoning subset and degrades the referring ones. This makes sense since the proposed alignment improves the reasoning abilities in complex and dynamic contexts. In the referring task, however, the target object is explicitly given in the text, and no complex reasoning is required. 

\begin{figure}[t]
\centering
\includegraphics[width=\linewidth]{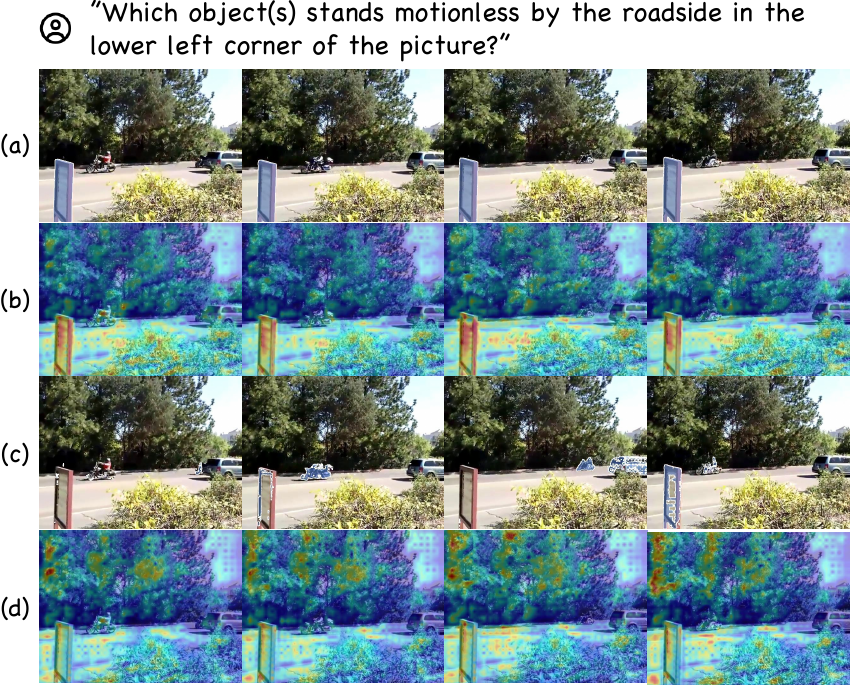}
   \caption{Visualization of attention in the MLLM and corresponding masks. (a) masks w/ Bi-Align. (b) attention w/ Bi-Align. (c) masks w/o Bi-Align. (d) attention w/o Bi-Align. }
\label{fig:cam}
\end{figure}

\begin{figure}[t]
\centering
\includegraphics[width=\linewidth]{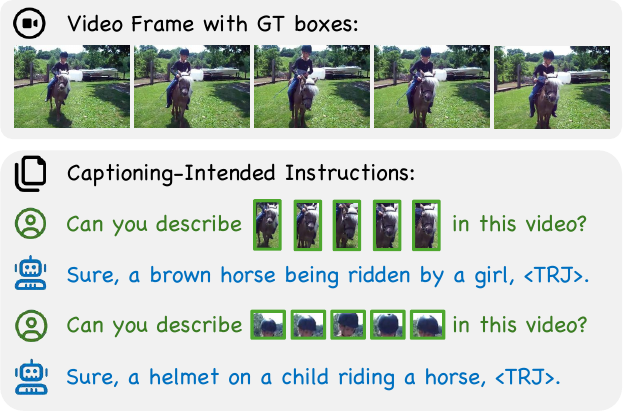}
   \caption{Visualization of captioning-intended instructions. }
\label{fig:caption}
\end{figure}

In Fig.~\ref{fig:caption}, we visualize the captioning results from our MLLM given the object trajectories. It is observed that our MLLM can well understand the context object trajectories. To probe the impact of bi-directional alignment on the MLLM, we visualize the attention maps in the MLLM in Fig.~\ref{fig:cam}. From the figure, it is clear that the bi-directional alignment brings better object perception by instructions, as well as better segmentation results. 

\newcommand{\jfwidthnum}{0.109\columnwidth}
\begin{table}[t!]
\footnotesize
\tabcolsep=0.057cm
\centering
\caption{Ablations on Bi-Align \& FCI. }
\label{tab:abl_both}
\vspace{-0.6em}
  \begin{tabular}{p{0.11\columnwidth}<{\centering} p{0.11\columnwidth}<{\centering}| p{\jfwidthnum}<{\centering}  p{\jfwidthnum}<{\centering} p{\jfwidthnum}<{\centering}| p{\jfwidthnum}<{\centering} p{\jfwidthnum}<{\centering} p{\jfwidthnum}<{\centering}}
  \toprule
     \multirow{2}{*}{Bi-Align} & \multirow{2}{*}{FCI} & \multicolumn{3}{c|}{ReVOS-referring} & \multicolumn{3}{c}{ReVOS-reasoning} \\
     \linevspace
     \cline{3-8}
     \linevspace
     & & 
     $\mathcal{J}\&\mathcal{F}$ & $\mathcal{J}$ & $\mathcal{F}$ & $\mathcal{J}\&\mathcal{F}$ & $\mathcal{J}$ & $\mathcal{F}$ \\
    \linevspace
    \hline 
    \textcolor[gray]{0.68}{\ding{55}} & \textcolor[gray]{0.68}{\ding{55}} & 53.0 & 50.9 & 55.1 & 42.8 & 41.0 & 44.6\\
    \ding{51} & \textcolor[gray]{0.68}{\ding{55}} & 54.4 & 52.7 & 56.1 & 43.5 & 41.3 & 45.6\\
    \textcolor[gray]{0.68}{\ding{55}} & \ding{51} & 53.5 & 51.6 & 55.4 & 43.2 & 41.2 & 45.2\\
    \rowcolor{blue!17!white}
    \ding{51} & \ding{51} & \textbf{55.0} & \textbf{53.1} & \textbf{57.0} & \textbf{45.5} & \textbf{43.5} & \textbf{47.4}\\
    \bottomrule
  \end{tabular}
\end{table}

\newcommand{\jfwidthrevosnf}{0.073\columnwidth}
\begin{table}[t!]
\footnotesize
\tabcolsep=0.1cm
\centering
\caption{{Ablations on Unified Mask Generator. KF: Number of key frames. BASE: Number of parameters of TrajSeg.} }
\label{tab:abl_num}
\vspace{-0.6em}
  \begin{tabular}{p{0.065\columnwidth}<{\centering} | 
  p{0.065\columnwidth}<{\centering} | p{0.21\columnwidth}<{\centering} | 
  p{\jfwidthrevosnf}<{\centering} p{\jfwidthrevosnf}<{\centering}
  p{\jfwidthrevosnf}<{\centering} | p{\jfwidthrevosnf}<{\centering}
  p{\jfwidthrevosnf}<{\centering} p{\jfwidthrevosnf}<{\centering}}
  \toprule
     \multirow{2}{*}{\scriptsize{Stage}} & \multirow{2}{*}{\scriptsize{KF}} & \multirow{2}{*}{\scriptsize{Params}} & \multicolumn{3}{c|}{ReVOS-referring} & \multicolumn{3}{c}{ReVOS-reasoning} \\
     \linevspace
     \cline{4-9}
     \linevspace
      & & & $\mathcal{J}\&\mathcal{F}$ & $\mathcal{J}$ & $\mathcal{F}$ & $\mathcal{J}\&\mathcal{F}$ & $\mathcal{J}$ & $\mathcal{F}$ \\
    \linevspace
    \hline
    \linevspace
    2 & 1 & BASE+224MB & 55.1 & 52.5 & 57.6 & 44.2 & 41.2 & 47.2 \\
    2 & 5 & BASE+224MB & \textbf{56.4} & 53.8 & \textbf{58.9} & 46.0 & 43.0 & \textbf{49.0}
 \\
    \linevspace
    \hline
    \linevspace
    1 & 1 & BASE & 55.7 & 53.2 & \underline{58.1} & \textbf{46.2} & 43.4 & \textbf{49.0} \\
    \rowcolor{blue!17!white}
    1 & 5 & BASE & \underline{56.0} & \textbf{53.9} & \underline{58.1} & \underline{46.1} & \textbf{43.7} & \underline{48.5}\\
    1 & 10 & BASE & 55.5 & 53.5 & 57.5 & 45.6 & 43.3 & 47.9\\
    \bottomrule
  \end{tabular}
\end{table}

\begin{table}[t!]
\footnotesize
\tabcolsep=0.06cm 
\centering
\caption{{Temporal robustness of unified mask generator on
varying KF. KF: Number of key frames.}}
\label{tab:kf_abl_final}
\vspace{-0.6em}
\begin{tabular}{
p{0.08\columnwidth}<{\centering} |
p{0.19\columnwidth}<{\centering} 
p{0.14\columnwidth}<{\centering} 
p{0.14\columnwidth}<{\centering}}
\toprule
\scriptsize{KF} &
\scriptsize{Avg-IoU$_{t\leftrightarrow t{+}1}$} &
\scriptsize{T-IoU-\!Var} &
\scriptsize{$\mathcal{J}\&\mathcal{F}$} \\
\hline
\rowcolor{white}
1 & \textbf{67.9} & \textbf{3.2} & 50.9 \\
\rowcolor{blue!17!white}
5 & 62.6 & 4.2 & \underline{\textbf{51.0}} \\
\rowcolor{white}
10 & 58.8 & 4.2 & 50.7 \\
\bottomrule
\end{tabular}
\end{table}

\begin{figure}[t]
\centering
\includegraphics[width=.99\linewidth]{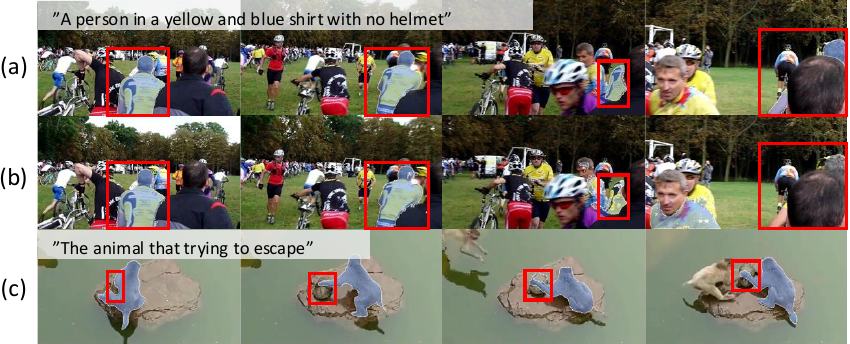}
\vspace{-0.9em}
   \caption{Qualitative results of ours w/ (a) and w/o (b) Unified Mask Generator. (c) Failure case. Red boxes are GT objects.}
\label{fig:fig_cam2}
\vspace{-1.1em}
\end{figure}

\paragraph{FCI Module} Tab.~\ref{tab:abl_both} verifies the effectiveness of the FCI module. We consider the same ablation setting as the bi-directional alignment. The table shows that TrajSeg w/ FCI works better than the one w/o FCI. This demonstrates that the FCI module unlocks the potential of the target token by transforming overall trajectory information into fine-grained spatial information.

\begin{figure}[t]
\centering
\includegraphics[width=\linewidth]{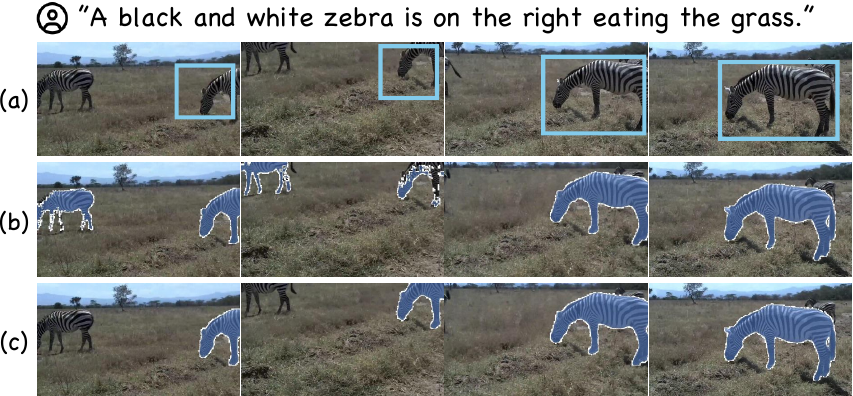}
   \caption{Qualitative ablations on the FCI module. (a) input video. (b) w/o FCI. (c) w/ FCI. }
\label{fig:fci}
\end{figure}

In addition, we provide several visualized examples to illustrate the impact of FCI on the key frame segmentation. As shown in Fig.~\ref{fig:fci}, the variant without FCI struggles on consistent object segmentation across frames. This can be mitigated by integrating frame-specific information into the target token, i.e., the FCI module. 

\begin{figure}[t]
\centering
\includegraphics[width=.85\linewidth]{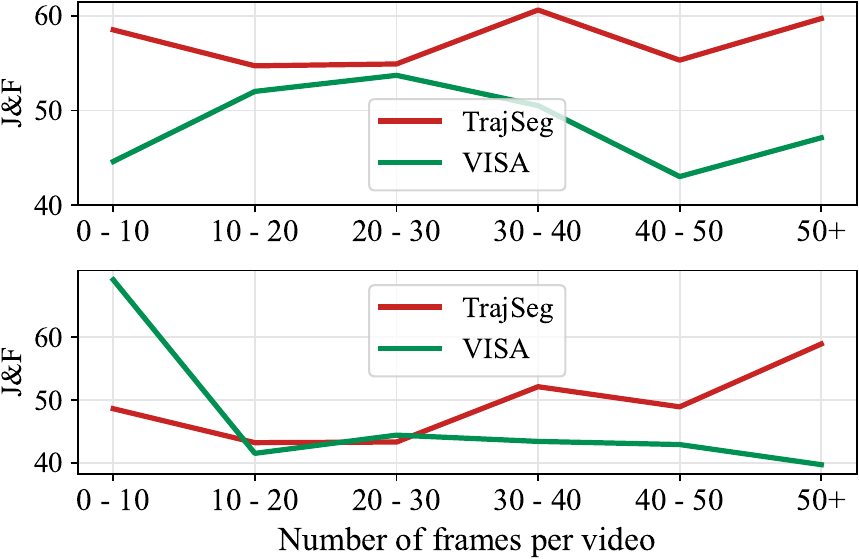}
   \caption{Segmentation scores on different video lengths. Top: ReVOS-Referring; Bottom: ReVOS-Reasoning. }
\label{fig:line}
\end{figure}

\paragraph{Number of Key Frames and Temporal Robustness} 
During inference, we randomly sample 1, 5, or 10 key frames to initialize and update memory. As shown in Tab.~\ref{tab:abl_num}, even using a single key frame already yields strong performance, demonstrating the effectiveness of the unified mask generator in capturing spatial-temporal relations. Compared to a two-stage pipeline (referring segmentation + visual tracker), our end-to-end design achieves comparable accuracy with significantly fewer parameters.

In addition, Fig.~\ref{fig:fig_cam2} and Tab.~\ref{tab:kf_abl_final} provide further evidence of temporal robustness. We evaluate temporal consistency using three complementary metrics: (1) IoU consistency between adjacent frames (Avg-IoU$_{t\leftrightarrow t+1}$), (2) variance of temporal IoU (T-IoU-Var), and (3) $\mathcal{J}\&\mathcal{F}$ for segmentation quality. Reducing the number of key frames improves temporal stability—Avg-IoU$_{t\leftrightarrow t+1}$ rises from 58.8 to 67.9 and T-IoU-Var drops from 4.2 to 3.2 when KF decreases from 10 to 1—indicating that the unified generator can effectively stabilize predictions over time. We finally adopt 5 key frames as a trade-off, achieving strong temporal robustness while maintaining high accuracy.

\paragraph{Long-term Reasoning Segmentation} 
Fig.~\ref{fig:line} compares TrajSeg and VISA on ReVOS videos with different lengths. It is clear that VISA cannot maintain high performance with the increase of video context. With trajectory-aware and unified mask generator, TrajSeg can understand and parse long-term object trajectories by instructions, achieving even better performance on long videos. 

\paragraph{Failure Case} 
As shown in Fig.~\ref{fig:fig_cam2}, the model fails when facing ambiguous instructions involving complex reasoning. In this case, multiple visually similar objects and the need to interpret nuanced human instructions make it difficult for the model to accurately ground the target, leading to incorrect mask localization. This reflects the current limitation of relying on static prompt alignment without deeper reasoning capability. Incorporating RLFT techniques such as GRPO may help enhance the model’s ability to handle such challenging scenarios.

\section{Conclusion}
\label{sec:conclusion}

This paper proposed TrajSeg, a trajectory-aware, and unified framework for video reasoning segmentation. With bidirectional text-trajectory alignment, TrajSeg achieves a better understanding of object trajectories and predicts instruction-relevant ones from videos. In addition, we proposed a novel FCI and a unified mask generator, enabling TrajSeg to have consistent predictions across frames in an end-to-end manner. Experimental results validate the effectiveness of TrajSeg. Despite achieving high performance on referring and reasoning segmentation, TrajSeg is limited by the quantity of sampled frames, hindering further improvement, and, therefore, calling for future solutions.

\section*{Acknowledgment}
This work was supported in part by the National Key R\&D Program of China under Grant 2022YFF1202903.

\bibliographystyle{IEEEtran}   
\bibliography{main}            

@String(CVPR= {IEEE Conf. Comput. Vis. Pattern Recog.})

@String(ICCV= {Int. Conf. Comput. Vis.})

@String(ECCV= {Eur. Conf. Comput. Vis.})

@String(ICLR = {Int. Conf. Learn. Represent.})

@String(AAAI = {AAAI})

@String(CVPR  = {CVPR})

@String(ICCV  = {ICCV})

@String(ECCV  = {ECCV})

@String(ICLR  = {ICLR})

@string(EMNLP = {EMNLP})

@inproceedings{seo2020urvos,
  title={Urvos: Unified referring video object segmentation network with a large-scale benchmark},
  author={Seo, Seonguk and Lee, Joon-Young and Han, Bohyung},
  booktitle={ECCV},
  year={2020}
}

@inproceedings{ding2023mevis,
  title={MeViS: A Large-scale Benchmark for Video Segmentation with Motion Expressions},
  author={Ding, Henghui and Liu, Chang and He, Shuting and Jiang, Xudong and Loy, Chen Change},
  booktitle={ICCV},
  year={2023}
}

@inproceedings{yan2024visa,
  title={VISA: Reasoning Video Object Segmentation via Large Language Models},
  author={Yan, Cilin and Wang, Haochen and Yan, Shilin and Jiang, Xiaolong and Hu, Yao and Kang, Guoliang and Xie, Weidi and Gavves, Efstratios},
  booktitle={ECCV},
  year={2024}
}

@inproceedings{yu2016modeling,
  title={Modeling context in referring expressions},
  author={Yu, Licheng and Poirson, Patrick and Yang, Shan and Berg, Alexander C and Berg, Tamara L},
  booktitle={ECCV},
  year={2016}
}

@inproceedings{mao2016generation,
  title={Generation and comprehension of unambiguous object descriptions},
  author={Mao, Junhua and Huang, Jonathan and Toshev, Alexander and Camburu, Oana and Yuille, Alan L and Murphy, Kevin},
  booktitle={CVPR},
  year={2016}
}

@inproceedings{liu2024visual,
  title={Visual instruction tuning},
  author={Liu, Haotian and Li, Chunyuan and Wu, Qingyang and Lee, Yong Jae},
  booktitle={NeurIPS},
  year={2024}
}

@inproceedings{han2024onellm,
  title={Onellm: One framework to align all modalities with language},
  author={Han, Jiaming and Gong, Kaixiong and Zhang, Yiyuan and Wang, Jiaqi and Zhang, Kaipeng and Lin, Dahua and Qiao, Yu and Gao, Peng and Yue, Xiangyu},
  booktitle={CVPR},
  year={2024}
}

@inproceedings{li2025llama,
  title={Llama-vid: An image is worth 2 tokens in large language models},
  author={Li, Yanwei and Wang, Chengyao and Jia, Jiaya},
  booktitle={ECCV},
  year={2024}
}

@article{yang2023improved,
  title={Lisa++: An improved baseline for reasoning segmentation with large language model},
  author={Yang, Senqiao and Qu, Tianyuan and Lai, Xin and Tian, Zhuotao and Peng, Bohao and Liu, Shu and Jia, Jiaya},
  journal={arXiv preprint arXiv:2312.17240},
  year={2023}
}

@inproceedings{pi2024perceptiongpt,
  title={Perceptiongpt: Effectively fusing visual perception into llm},
  author={Pi, Renjie and Yao, Lewei and Gao, Jiahui and Zhang, Jipeng and Zhang, Tong},
  booktitle={CVPR},
  year={2024}
}

@inproceedings{ren2024pixellm,
  title={Pixellm: Pixel reasoning with large multimodal model},
  author={Ren, Zhongwei and Huang, Zhicheng and Wei, Yunchao and Zhao, Yao and Fu, Dongmei and Feng, Jiashi and Jin, Xiaojie},
  booktitle={CVPR},
  year={2024}
}

@inproceedings{he2024multi,
  title={Multi-modal Instruction Tuned LLMs with Fine-grained Visual Perception},
  author={He, Junwen and Wang, Yifan and Wang, Lijun and Lu, Huchuan and He, Jun-Yan and Lan, Jin-Peng and Luo, Bin and Xie, Xuansong},
  booktitle={CVPR},
  year={2024}
}

@inproceedings{xia2024gsva,
  title={Gsva: Generalized segmentation via multimodal large language models},
  author={Xia, Zhuofan and Han, Dongchen and Han, Yizeng and Pan, Xuran and Song, Shiji and Huang, Gao},
  booktitle={CVPR},
  year={2024}
}

@inproceedings{bai2024one,
  title={One Token to Seg Them All: Language Instructed Reasoning Segmentation in Videos},
  author={Bai, Zechen and He, Tong and Mei, Haiyang and Wang, Pichao and Gao, Ziteng and Chen, Joya and Liu, Lei and Zhang, Zheng and Shou, Mike Zheng},
  booktitle={NeurIPS},
  year={2024}
}

@inproceedings{lai2024lisa,
  title={Lisa: Reasoning segmentation via large language model},
  author={Lai, Xin and Tian, Zhuotao and Chen, Yukang and Li, Yanwei and Yuan, Yuhui and Liu, Shu and Jia, Jiaya},
  booktitle={CVPR},
  year={2024}
}

@inproceedings{wu2022language,
  title={Language as queries for referring video object segmentation},
  author={Wu, Jiannan and Jiang, Yi and Sun, Peize and Yuan, Zehuan and Luo, Ping},
  booktitle={CVPR},
  year={2022}
}

@inproceedings{botach2022end,
  title={End-to-end referring video object segmentation with multimodal transformers},
  author={Botach, Adam and Zheltonozhskii, Evgenii and Baskin, Chaim},
  booktitle={CVPR},
  year={2022}
}

@inproceedings{he2024decoupling,
  title={Decoupling Static and Hierarchical Motion Perception for Referring Video Segmentation},
  author={He, Shuting and Ding, Henghui},
  booktitle={CVPR},
  year={2024}
}

@inproceedings{han2023html,
  title={Html: Hybrid temporal-scale multimodal learning framework for referring video object segmentation},
  author={Han, Mingfei and Wang, Yali and Li, Zhihui and Yao, Lina and Chang, Xiaojun and Qiao, Yu},
  booktitle={ICCV},
  year={2023}
}

@inproceedings{yan2023referred,
  title={Referred by Multi-Modality: A Unified Temporal Transformer for Video Object Segmentation},
  author={Yan, Shilin and Zhang, Renrui and Guo, Ziyu and Chen, Wenchao and Zhang, Wei and Li, Hongyang and Qiao, Yu and He, Zhongjiang and Gao, Peng},
  booktitle={AAAI},
  year={2024}
}

@inproceedings{luo2023soc,
  title={SOC: Semantic-Assisted Object Cluster for Referring Video Object Segmentation},
  author={Luo, Zhuoyan and Xiao, Yicheng and Liu, Yong and Li, Shuyan and Wang, Yitong and Tang, Yansong and Li, Xiu and Yang, Yujiu},
  booktitle={NeurIPS},
  year={2023}
}

@inproceedings{miao2023spectrum,
  title={Spectrum-guided Multi-granularity Referring Video Object Segmentation},
  author={Miao, Bo and Bennamoun, Mohammed and Gao, Yongsheng and Mian, Ajmal},
  booktitle={ICCV},
  year={2023}
}

@inproceedings{zhu2024exploring,
  title={Exploring Pre-trained Text-to-Video Diffusion Models for Referring Video Object Segmentation},
  author={Zhu, Zixin and Feng, Xuelu and Chen, Dongdong and Yuan, Junsong and Qiao, Chunming and Hua, Gang},
  booktitle={ECCV},
  year={2024}
}

@inproceedings{yuan2024losh,
  title={Losh: Long-short text joint prediction network for referring video object segmentation},
  author={Yuan, Linfeng and Shi, Miaojing and Yue, Zijie and Chen, Qijun},
  booktitle={CVPR},
  year={2024}
}

@article{trackgpt,
  title={Tracking with Human-Intent Reasoning},
  author={Zhu, Jiawen and Cheng, Zhi-Qi and He, Jun-Yan and Li, Chenyang and Luo, Bin and Lu, Huchuan and Geng, Yifeng and Xie, Xuansong},
  journal={arXiv preprint arXiv:2312.17448},
  year={2023}
}

@inproceedings{wu2023onlinerefer,
  title={OnlineRefer: A Simple Online Baseline for Referring Video Object Segmentation},
  author={Wu, Dongming and Wang, Tiancai and Zhang, Yuang and Zhang, Xiangyu and Shen, Jianbing},
  booktitle={ICCV},
  year={2023}
}

@inproceedings{cheng2022xmem,
  title={{XMem}: Long-Term Video Object Segmentation with an Atkinson-Shiffrin Memory Model},
  author={Cheng, Ho Kei and Schwing, Alexander G.},
  booktitle={ECCV},
  year={2022}
}

@inproceedings{bekuzarov2023xmem++,
  title={Xmem++: Production-level video segmentation from few annotated frames},
  author={Bekuzarov, Maksym and Bermudez, Ariana and Lee, Joon-Young and Li, Hao},
  booktitle={ICCV},
  year={2023}
}

@inproceedings{zhou2017scene,
  title={Scene parsing through ade20k dataset},
  author={Zhou, Bolei and Zhao, Hang and Puig, Xavier and Fidler, Sanja and Barriuso, Adela and Torralba, Antonio},
  booktitle={CVPR},
  year={2017}
}

@inproceedings{caesar2018coco,
  title={Coco-stuff: Thing and stuff classes in context},
  author={Caesar, Holger and Uijlings, Jasper and Ferrari, Vittorio},
  booktitle={CVPR},
  year={2018}
}

@inproceedings{ramanathan2023paco,
  title={{PACO}: Parts and Attributes of Common Objects},
  author={Ramanathan, Vignesh and Kalia, Anmol and Petrovic, Vladan and Wen, Yi and Zheng, Baixue and Guo, Baishan and Wang, Rui and Marquez, Aaron and Kovvuri, Rama and Kadian, Abhishek and others},
  booktitle={CVPR},
  year={2023}
}

@inproceedings{chen2014detect,
  title={Detect what you can: Detecting and representing objects using holistic models and body parts},
  author={Chen, Xianjie and Mottaghi, Roozbeh and Liu, Xiaobai and Fidler, Sanja and Urtasun, Raquel and Yuille, Alan},
  booktitle={CVPR},
  year={2014}
}

@inproceedings{liu2023improvedllava,
  title={Improved Baselines with Visual Instruction Tuning},
  author={Liu, Haotian and Li, Chunyuan and Li, Yuheng and Lee, Yong Jae},
  booktitle={CVPR},
  year={2024}
}

@inproceedings{kazemzade2014referring,
  title={Referring to objects in photographs of natural scenes},
  author={Kazemzadeh, Sahar and Ordonez, Vicente and Matten, Mark and Berg, Tamara},
  booktitle={EMNLP},
  year={2014}
}

@inproceedings{ravi2024sam,
  title={{SAM 2}: Segment Anything in Images and Videos},
  author={Ravi, Nikhila and Gabeur, Valentin and Hu, Yuan-Ting and Hu, Ronghang and Ryali, Chaitanya and Ma, Tengyu and Khedr, Haitham and R{\"a}dle, Roman and Rolland, Chloe and Gustafson, Laura and others},
  booktitle={ICLR},
  year={2024}
}

@inproceedings{hu2021lora,
  title={{LoRA}: Low-Rank Adaptation of Large Language Models},
  author={Hu, Edward J. and Shen, Yelong and Wallis, Phillip and Allen-Zhu, Zeyuan and Li, Yuanzhi and Wang, Shean and Wang, Liang and Chen, Weizhu and others},
  booktitle={ICLR},
  year={2022}
}

@inproceedings{rasley2020deepspeed,
  title={Deepspeed: System optimizations enable training deep learning models with over 100 billion parameters},
  author={Rasley, Jeff and Rajbhandari, Samyam and Ruwase, Olatunji and He, Yuxiong},
  booktitle={SIGKDD},
  year={2020}
}

@inproceedings{loshchilov2017decoupled,
  title={Decoupled Weight Decay Regularization},
  author={Loshchilov, Ilya and Hutter, Frank},
  booktitle={ICLR},
  year={2017}
}

@article{yin2023survey,
  title={A Survey on Multimodal Large Language Models},
  author={Yin, Shukang and Fu, Chaoyou and Zhao, Sirui and Li, Ke and Sun, Xing and Xu, Tong and Chen, Enhong},
  journal={National Science Review},
  volume={11},
  number={12},
  year={2024},
  publisher={Oxford University Press}
}

@inproceedings{kirillov2023segment,
  title={Segment anything},
  author={Kirillov, Alexander and Mintun, Eric and Ravi, Nikhila and Mao, Hanzi and Rolland, Chloe and Gustafson, Laura and Xiao, Tete and Whitehead, Spencer and Berg, Alexander C and Lo, Wan-Yen and others},
  booktitle={ICCV},
  year={2023}
}

@inproceedings{liang2025referdino,
  title={Referdino: Referring video object segmentation with visual grounding foundations},
  author={Liang, Tianming and Lin, Kun-Yu and Tan, Chaolei and Zhang, Jianguo and Zheng, Wei-Shi and Hu, Jian-Fang},
  booktitle={ICCV},
  year={2025}
}

@inproceedings{pan2025semantic,
  title={Semantic and sequential alignment for referring video object segmentation},
  author={Pan, Feiyu and Fang, Hao and Li, Fangkai and Xu, Yanyu and Li, Yawei and Benini, Luca and Lu, Xiankai},
  booktitle={CVPR},
  year={2025}
}

@inproceedings{huang2025unleashing,
  title={Unleashing the temporal-spatial reasoning capacity of gpt for training-free audio and language referenced video object segmentation},
  author={Huang, Shaofei and Ling, Rui and Li, Hongyu and Hui, Tianrui and Tang, Zongheng and Wei, Xiaoming and Han, Jizhong and Liu, Si},
  booktitle={AAAI},
  year={2025}
}

\vfill

\end{document}